%% file: 0_main.tex
\pdfoutput=1
\documentclass[11pt]{article}

\usepackage[final]{acl}

\usepackage{times}
\usepackage{latexsym}

\usepackage[T1]{fontenc}
\usepackage[utf8]{inputenc}

\usepackage{microtype}

\usepackage{inconsolata}

\usepackage{graphicx}

\usepackage{enumitem}
\usepackage{graphicx}
\usepackage{quoting}
\usepackage{colortbl}
\usepackage{tabularx}
\usepackage{tabulary}
\usepackage{siunitx}
\usepackage{listings}
\usepackage{comment}
\usepackage{booktabs}
\usepackage{multicol, multirow, tabulary, siunitx, bm}
\usepackage{placeins}
\usepackage{subcaption}
\usepackage{amssymb} 
\usepackage{sansmath}
\usepackage[framemethod=tikz]{mdframed}

\newmdenv[
  backgroundcolor=yellow!15,
  linecolor=black,
  linewidth=0.3pt,
  roundcorner=1pt,
  skipabove=1em,
  skipbelow=1em,
  innertopmargin=4pt,
  innerbottommargin=4pt,
  innerleftmargin=4pt,
  innerrightmargin=4pt,
]{keyinsight}

\usepackage{xcolor}

\usepackage{layout}

\setlist[description]{style=sameline, leftmargin=1em}

\title{Certain but not Probable? Differentiating Certainty from Probability in LLM Token Outputs for Probabilistic Scenarios}

\author{Autumn Toney-Wails \\
  SciTech Strategies, Inc. \\
  Georgetown University  \\
  \texttt{autumn@mapofscience.com} \\\And
  Ryan Wails \\ Georgetown University \\
  \texttt{rsw66@georgetown.edu}}

\begin{document}
\maketitle
\begin{abstract}
Reliable uncertainty quantification (UQ) is essential for ensuring trustworthy downstream use of large language models, especially when they are deployed in decision-support and other knowledge-intensive applications. Model certainty can be estimated from token logits, with derived probability and entropy values offering insight into performance on the prompt task. However, this approach may be inadequate for probabilistic scenarios, where the probabilities of token outputs are expected to align with the theoretical probabilities of the possible outcomes. We investigate the relationship between token certainty and alignment with theoretical probability distributions in well-defined probabilistic scenarios. Using GPT-4.1 and DeepSeek-Chat, we evaluate model responses to ten prompts involving probability (e.g., roll a six-sided die), both with and without explicit probability cues in the prompt (e.g., roll a \textit{fair} six-sided die). We measure two dimensions: (1) response validity with respect to scenario constraints, and (2) alignment between token-level output probabilities and theoretical probabilities. Our results indicate that, while both models achieve perfect in-domain response accuracy across all prompt scenarios, their token-level probability and entropy values consistently diverge from the corresponding theoretical distributions. 
\end{abstract}

\input{latex/1_introduction}
\input{latex/2_relatedwork}
%\input{latex/3_definitions}
\input{latex/4_methodology}
\input{latex/5_experiments}
\input{latex/6_discussion}

\input{latex/7_conclusion}
\bibliography{custom}

 %\newpage
 %\layout
 
 %\makeatletter
 %orig: \f@size
 %\verb+\tiny+ \tiny \f@size
 %\verb+\scriptsize+ \scriptsize \f@size
 %\verb+\footnotesize+ \footnotesize \f@size
 %\verb+\small+ \small \f@size
 %\verb+\normalsize+ \normalsize \f@size
 %verb+\large+ \large \f@size
 %\verb+\Huge+ \Huge \f@size
 %\makeatother

\end{document}

%% file: latex/1_introduction.tex
\section{Introduction}
\label{sec:Intro}

As large language models (LLMs) are increasingly integrated into decision-support and knowledge-intensive applications, uncertainty quantification (UQ) is essential to ensure reliable downstream use \cite{xiong2024efficient,vashurin2025benchmarking}. Prior work has focused on leveraging the token logits---numerical representations encoding the model’s output probabilities---for UQ methods applied to natural language generation tasks \cite{malinin2020uncertainty,kuhn2023semantic,gupta2024language,lin-etal-2024-contextualized,duan-etal-2024-shifting,fadeeva-etal-2024-fact,lovering-etal-2025-language}. Token logits can be transformed into probabilities using activation functions (e.g., softmax or sigmoid), enabling entropy computation over the token distribution. These probability and entropy values are often used to quantify model certainty, providing token-level or response-level confidence scores to users. 

For prompts involving randomness, risk, or chance, traditional UQ alone may be insufficient to capture the true confidence a user should have in implementing a model.
\begin{keyinsight}
\emph{Key insight:}
Specifically, for prompts involving probabilistic scenarios with inherent aleatoric uncertainty (e.g., ``flip a fair coin''), a model's behavior is trustworthy only when its
distribution over possible outputs matches the intended theoretical distribution (which may be only implicitly specified). 
\end{keyinsight}
Hence, in these probabilistic scenarios, we argue there are two considerations for uncertainty quantification that are particularly important:
\begin{enumerate}
    \item Whether a response is a valid output under the specified scenario constraints (e.g., if the prompt is ``roll a six-sided dice'', the output response should not be ``7'').
    \item Whether the response probability aligns with the underlying theoretical distribution. 
\end{enumerate}
These considerations are somewhat in contrast to traditional UQ settings, where model accuracy typically corresponds to low uncertainty when a model is well-calibrated.

\begin{figure}[t!]
    \centering
    \includegraphics[width=1.0\columnwidth]{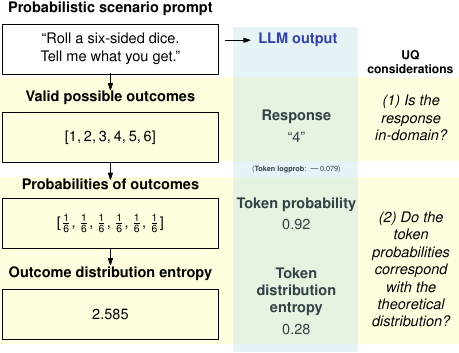}
    \caption{Probabilistic scenario prompting and response evaluation design.}
    \label{fig:pipeline}
\end{figure}

In this work, we explore the relationship between token certainty and theoretical probability in well-defined probabilistic scenarios. We prompt GPT-4.1 and DeepSeek with 10 well-known scenarios (e.g., roll a six-sided die or pick a card from a deck of 52 cards). Our selected scenarios allow us to measure the models' response certainty (was the response valid?) and to compare the token output probability and entropy of possible output tokens to theoretical distributions (does the response probability align to real-world distributions?). We provide an example of our experimental design framework in Figure \ref{fig:pipeline}.  We experiment with a second series of prompting, in which we include language that reflects the distribution the model should use (e.g., roll a \textit{fair} six-sided die, pick a card from a deck of 52 cards \textit{uniformly at random}).

Our findings suggest that although GPT-4.1 and DeepSeek exhibit appropriate contextual understanding and high response certainty, their token-level output probabilities do not reliably represent true probabilities in scenarios that require statistical reasoning or random sampling. Specifically, we find that they respond with valid outputs with 100\% accuracy (i.e., they understand the constraints of the prompted scenario) but their probability and entropy values never align with the corresponding theoretical values. Based on our findings, we explore three research questions for further discussion:
\begin{description}
    \item[RQ1] Can LLMs accurately reason about the theoretical probabilities of our prompt scenarios?
    \item[RQ2] Are LLMs appropriate and reliable tools for probability-oriented tasks where usability depends on alignment with theoretical distributions?
    \item[RQ3] How can uncertainty quantification methods be adapted to jointly evaluate response validity and distribution alignment for probability-driven tasks?
\end{description}

%% file: latex/2_relatedwork.tex
\section{Background and Related Work}

\subsection{Uncertainty Quantification for LLMs}
Uncertainty quantification (UQ) measures the confidence associated with model output, providing information to guide user decisions on model deployment, refinement, or rejection. UQ for LLMs spans a wide range of methodologies for varying granularity levels and certainty dimensions \cite{liu2025uncertainty,shorinwa2025survey}. Depending on the task, UQ can be computed at the token level (a single word), the span level (a contiguous sequence of tokens representing the response segment of interest), or the response level (the entire generated output). To organize the wide range of UQ methods, \citet{liu2025uncertainty} and \citet{shorinwa2025survey} present surveys that synthesize the related literature and characterize the state of the field. 

\citet{liu2025uncertainty} proposed a taxonomy to organize sources of uncertainty beyond the general aleatoric (uncertainty stemming from the randomness or variability in the data) and epistemic (uncertainty stemming from lack of knowledge) categorization. The authors provide four dimensions of uncertainty for finer-grained analysis: (1) input uncertainty, (2) reasoning uncertainty, (3) parameter uncertainty, and (4) prediction uncertainty. This taxonomy supports their survey design, which focuses on the relationships between model scale, open-ended generation, and uncertainty dynamics. The authors further outline research directions, emphasizing the need for enhanced UQ methods for natural language generation (NLG) tasks that extend beyond traditional binary formulations.

\citet{shorinwa2025survey} provide an in-depth overview on UQ, starting with the application in machine learning tasks before focusing on applications in LLMs. The authors frame their survey around the characteristics of the transformer architecture and the auto-regressive, token-by-token generation process underlying NLG.  The survey is organized by UQ methods: (1) token-level, (2) self-verbalized, (3) semantic-similarity, and (4) mechanistic interpretability. In conclusion, Shorinwa et al. provide 5  directions for future work that recognize the common mistakes and unique characteristics of UQ for LLM-generated output (e.g., distinguishing consistency from factuality and recognizing that entropy does not equate to factuality).

\subsection{Task-Dependent UQ}

The choice of UQ granularity and methodology is task-dependent---relevant certainty dimensions should be determined by the user’s objectives and appropriately inform reasoning about model outputs. Prior work has investigated methods to measure certainty based on semantic similarity \cite{kuhn2023semantic}, fact-checking information claims \cite{fadeeva-etal-2024-fact}, and the calibration of probability distributions \cite{lovering-etal-2025-language}. These certainty dimensions provide user insight into model performance on common aspects of generated output; for example, whether multiple valid phrasings exist for an idea, whether model confidence reflects factual accuracy, or whether responses to probabilistic scenarios align with formally defined theoretical distributions. 

\citet{kuhn2023semantic} defined semantic entropy, a UQ measurement that captures semantic meaning, to improve predictive model accuracy on question and answering (QA) tasks. Using the GPT-like OPT models \cite{zhang2022opt}, the authors experiment with the TriviaQA \cite{joshi-etal-2017-triviaqa} and CoQA \cite{reddy2019coqa} datasets. The semantic entropy measurements outperformed baseline measurements by calculating the entropy of the distribution over meanings rather than token sequence. 

\citet{fadeeva-etal-2024-fact} proposed a claim-specific certainty method, Claim Conditioned Probability, to identify the factual accuracy of the generated claim and compare it to the model's response confidence. This technique allows for efficient hallucination detection and provides the end user with a measurement of model certainty about the specific information claim, as opposed to the overall response certainty. Experimenting with English (Vicuna 13B, Mistral 7B, Jais 13B, and GPT-3.5-turbo), Chinese (Yi 6B), Arabic (Jais 13B and GPT-4), and Russian (Vikhr-instruct-0.2 7B),
the authors evaluated their method using human annotation and found that their results are comparable to fact-checking efforts using external knowledge sources.

\citet{lovering-etal-2025-language} investigated if LLM generated output probabilities are calibrated
to the underlying defined probabilities within their textual contexts. Using a set of word problems that define the probabilistic scenario (e.g., ``From 17 red marbles and 99 blue marbles, Billy reached blindly into the bag and
grabbed a marble with the color [red/blue].''), the authors prompted Mistral, Yi, Gemma, and GPT-4 model families to generate outputs and associated token probabilities. Lovering et al. found that these GPT LLMs are sensitive to the input prompt and do not produce outputs that are calibrated to the presented distributions.

We build on the research presented by \citet{lovering-etal-2025-language} and incorporate the veracity dimension from \citet{fadeeva-etal-2024-fact} by prompting GPT models with well-known probabilistic scenarios. Our experimental design mitigates the prompt-induced bias (found in Lovering et al.'s study) by omitting explicit specification of the probabilistic scenario. Thus, we include a response validation to ensure that the generated outputs are valid within the scenario's constraints.

%Because UQ can measure specific dimensions of LLM outputs, there is not a universal approach for any given task.

%Additionally, these studies highlight the importance of selecting task-relevant UQ methods and support holistic approaches that encompass the users implementation objectives. 

%These specific certainty dimensions provide user insight to model performance on common aspects of generated output: are there many ways to express an idea? Does the model confidence reflect factual accuracy? Was the response to a statistical scenario aligned to the defined, real-world distributions? These nuances are important when thinking about response evaluation wholistically as opposed to simply low entropy over the token distribution means high confidence, and therefore I should trust the model and use it in my system.

%% file: latex/4_methodology.tex
\section{Methodology}
\label{sec:methods}

\subsection{Definitions}

This work is primarily focused on two measurements of LLM outputs:

\begin{description}
    \item[Token probability] Each token $t \in \mathcal{T}$ has a probability of being selected by the model as output, $0 \leq \Pr(t) \leq 1$.
    The probabilities of all possible tokens sums to one: $\sum_{t \in \mathcal{T}} \Pr(t) = 1$.
    Language model token probabilities are conditioned on previous tokens. In other words, the probability of the $n$th token output by the model is $\Pr(t_n \mid t_{n-1} t_{n-2} \dots t_k)$, where previous tokens may come from user dialogue or previous model outputs.
    \item[Entropy] The entropy of a discrete random variable $T$ is defined as $$\mathrm{H}(T) = -\sum_{t \in \mathcal{T}}\log_2(\Pr(t))\cdot\Pr(t)$$ where $T$ takes on values in the set $\mathcal{T}$. Entropy quantifies how much uncertainty is associated with the variable; if a variable has low entropy, then its outcome is easily predicted.
\end{description}

\subsection{Computation}

\paragraph{Token probability}
Certain LLM vendors, such as OpenAI and DeepSeek via the \textit{Chat Completions} API, allow token ``logprobs'' to be programmatically accessed by a user. For a token $t$, its logprob is
$l(t) = \log\left(\Pr\left(t\right)\right)$; hence, token probability can be easily derived, $e^{l(t)}~=~\Pr\left(t\right)$.

\paragraph{Entropy}
Given a set of token (log-)probabilities from the model, entropy can be computed as described above. Note that there are two limitations that affect this procedure. First, vendors expose only a limited number of token logprobs (at the time of this writing, OpenAI and DeepSeek expose the top 20 most probable tokens and their logprobs). To obtain a proper probability distribution, we add an {\tt other} token to the collection with a determined probability value so that the probability values sum to 1. Second, some LLM output values are the concatenation of multiple tokens. Token logprobs, however, are provided for only the \emph{selected} sequence of tokens and not for all possible sequences. For entropy to be computed correctly, the distribution should be taken over all possible output sequences. For both of these reasons, computed entropy values are likely a slight underestimate of the model's true entropy over the token distribution.

%\subsection{LLM Selection}
%We experiment with two state-of-the-art LLMs: OpenAI's GPT-4.1 (\texttt{gpt-4.1-2025-04-14}) and DeepSeek-Chat (\texttt{DeepSeek-R1-0528}). Both models are closed-source implementations of mixture-of-experts architecture, accessible via API endpoints. We selected these models on three criteria: (1) competitive benchmark performance across a range of natural language generation tasks, (2) architectural features representative of current frontier LLM designs, and (3) explicit support for token-level log probability (\texttt{logprobs}) outputs, which are required for our experimental analysis. 

\subsection{Prompt Design and LLM Configuration}
We select 10 prompts that contain well-defined probabilistic scenarios; five scenarios are various actions of chance, and five scenarios are random choices from a set of items. For further evaluation, we include a second prompt set that explicitly instructs how we want the model to select an output (e.g., flip a \textit{fair} coin, pick a Shakespeare play \textit{uniformly at random}). We list the scenarios and note the terminology included in their specified versions in Table \ref{tab:prompts} with their corresponding outcome probability and entropy values. Additionally, we include the statement ``Respond only with the result'' to the end of all prompts, as the generated output affects the token probabilities. We want only the exact response to be generated for our evaluations.

\begin{table*}[ht]
    \small
    \sffamily
    \centering
    \setlength{\tabcolsep}{5pt}
    \begin{tabulary}{\textwidth}{LLS[table-format=1.4]S[table-format=1.3]}
         \toprule
          \textbf{Scenario} & \textbf{Prompt} & \textbf{$\bm{\mathsf {Pr(x)}}$} & \textbf{$\bm{\mathsf{H(X)}}$}   \\ \midrule

        Bible\textsuperscript{\textdagger} & Pick a book of the Bible. & 0.0152 & 6.04 \\
        
       Bingo\textsuperscript{\textdagger} & Pick a bingo ball. Tell me what you get. & 0.0133 & 6.23\\
       
        Coin flip\textsuperscript{\(\diamondsuit\)} & Flip a coin. & 0.5 & 1 \\
        Dart on Map\textsuperscript{*} & Throw a dart at a map. Tell me what country it lands on. & 0.00403 & 7.96 \\
        Die roll\textsuperscript{\(\diamondsuit\)} & Roll a six-sided die. Tell me what you get. & 0.167 & 2.58 \\
        Month \& Day\textsuperscript{\textdagger} & Pick a month and day. & 0.0027 & 8.51\\
        Playing Cards\textsuperscript{\textdagger} & Pick a card from a deck of playing cards. Tell me your card. & 0.0192 & 5.7\\
        R-P-S\textsuperscript{\textdagger} & You are playing rock, paper, scissors. Make your throw. & 0.33 & 1.59 \\
        Roulette\textsuperscript{*} & Spin an American roulette wheel. Tell me which pocket it lands in. & 0.0263 & 5.25 \\
        Shakespeare\textsuperscript{\textdagger} & Pick a Shakespeare play. & 0.0256 & 5.29\\
        
    \bottomrule
    \end{tabulary}
    \caption{Probabilistic scenarios used for prompt experiments with their corresponding outcome probability and entropy values under a uniform distribution. \textsuperscript{\textdagger} denotes the specified scenarios that add ``uniformly at random'', \textsuperscript{*} denotes the specified scenarios that add ``randomly'', and \textsuperscript{\(\diamondsuit\)} denotes the specified scenarios that add ``fair''.}
\label{tab:prompts}
\end{table*}

We experiment with two state-of-the-art LLMs: OpenAI's GPT-4.1 (\texttt{gpt-4.1-2025-04-14}) and DeepSeek-Chat (\texttt{DeepSeek-R1-0528}). Both models are closed-source implementations accessible via API endpoints. We selected these models on three criteria: (1) competitive benchmark performance across a range of natural language generation tasks, (2) architectural features representative of current frontier LLM designs, and (3) explicit support for token-level logprob outputs, which are required for our experimental analysis. 

We prompt GPT-4.1 and DeepSeek-Chat via the \texttt{OpenAI} python package using \texttt{chat.completions} and set \texttt{logprobs} $=$ \texttt{true} to output the maximum (20) top tokens and their corresponding logprob values. Our prompts are passed through the user role and we do not assign a system role. We use the default values for the remaining parameters.
Our code is available at \url{https://github.com/autumntoney/chatbot-certainty}.

%% file: latex/5_experiments.tex
\section{Experimental Results}
For both the unspecified and specified prompt sets, we generate five independent samples from each model. We compute the mean token-level probability and entropy values across these samples and compare the aggregated statistics to the theoretical values defined by the corresponding probabilistic scenario (listed in Table \ref{tab:prompts}). Table \ref{tab:top_responses} displays 
 the most frequent responses for each experiment configuration in which at least one response was repeated across samples. Configurations in which all responses were unique are not listed. The frequency column denotes the total count of response occurrences across both sets (e.g., if the model generates the same response three times in the unspecified prompt and four times in the specified prompt, then the frequency value is seven). 

\newcommand{\repeatdash}{\multicolumn{1}{l}{\textemdash}}
\newcommand{\na}{\multicolumn{1}{l}{\textsc{n/a}}}
\newcommand{\nac}{\multicolumn{1}{c}{\textsc{n/a}}}

\begin{table*}[t]
    \sffamily
	\centering
	\small
	\begin{tabulary}{\textwidth}{LLLLS[table-format=2]S[table-format=2]}
		\toprule
		\multirow{2}{*}{\bf \sffamily Scenario} & \multirow{2}{*}{\bf\sffamily Dist} & \multicolumn{2}{c}{\bf\sffamily Response} & \multicolumn{2}{c}{\bf\sffamily Frequency} \\
		\cmidrule(lr){3-4} \cmidrule(lr){5-6}
		& & {\bf\sffamily DeepSeek} & {\bf\sffamily GPT-4.1} & {\bf\sffamily DeepSeek} &
		{\bf\sffamily GPT-4.1}  \\ \midrule

		Coin flip & U/S & Heads & Heads & 10 & 10 \\
		Die roll & U/S & 4 & 4 & 10 & 10 \\
		Rock-paper-scissors & U/S & Scissors & Scissors & 9 & 6 \\

		Roulette & U  & 14 & 27 Black & 2 & 2 \\
		\repeatdash & S & 14 \& 17 & 32 Red & 4 & 2 \\

		Dart on Map & U  & Botswana & Uzbekistan & 4 & 4 \\
	  \repeatdash & S & Mongolia & Brazil & 4 & 2\\

		Playing Cards & U/S  & 7 of Hearts & Queen of Hearts & 8 & 6 \\

		Shakespeare & U & Hamlet & Macbeth & 5 & 4 \\
		\repeatdash & S & King Lear & Twelfth Night & 2 & 3\\

		Month \& Day & U & July 12 \& 15 & \na & 4 & \nac \\
		\repeatdash & S & June 14 & \na & 3 & \nac \\

		Bingo & U & 17 & \na & 5 & \nac \\
		\repeatdash & S & 42 & \na & 4 & \nac \\
		Bible & U & Genesis & Ruth & 3 & 3 \\
		\repeatdash & S & Jonah & Habakkuk & 3 & 2 \\
		\bottomrule
	\end{tabulary}
	\caption{Most frequent responses by prompt scenario and model. Each row
  reports results where the distribution was explicitly specified (S) or
  unspecified(U); U/S indicates that the results were the same for both the
  specified and unspecified cases. If a model never generated a token more than
  once, the columns are marked \textsc{n/a}.}
	\label{tab:top_responses}
\end{table*}

We find that the majority of samples have more than one response in common except for GPT-4.1's Bingo (unspecified and specified) and  month and day (specified) samples. For the coin-flip and dice-roll scenarios, both DeepSeek-Chat and GPT-4.1 produce the same outcome across all samples (“heads” and “4”, respectively), and in the rock–paper–scissors scenario, both models predominantly select “scissors.” The most frequent responses differ depending on the model and prompt for all other scenarios.

All responses from both models are within the probabilistic scenario constraints, with the exception of DeepSeek-Chat providing partial responses for Bingo and Roulette. For example, GPT-4.1 specifies the roulette pocket with both color and number (e.g., 27 Black), whereas DeepSeek-Chat provides only the number (e.g., 14). Similarly, in the Bingo scenario, GPT-4.1 provides the letter and number (e.g., G-52) but DeepSeek only provides a number (e.g., 17). We consider these partial responses to be accurate within the scenario as the numeric values are valid options.

\subsection{Prompt-level Comparisons}

To compare the generated outputs between the unspecified and specified prompts, we compute the differences between their mean probability and entropy values across the five samples for each prompt set. Table~\ref{table:results} displays the complete set of probability and entropy values, with the differences between results obtained from unspecified versus specified prompts. For all unspecified prompt results, both DeepSeek-Chat and GPT-4.1 produce probability values that are higher and entropy values that are lower than their corresponding theoretical values. Thus, we expect that the results generated from the specified prompts will decrease the probability values and increase the entropy values. We find that while the probability and entropy are slightly corrected with explicitly-specified prompts, the rate of correction is inconsistent and the specified values remain far from the correct theoretical values.

%\autumn{@Ryan should we say something like we expect to see the values shift in a way more towards the theoretical values which would mean decreasing probability and increasing entropy?}

\sisetup{separate-uncertainty = false}

\begin{table*}[t]
    \centering
	\footnotesize
    \sffamily
    \setlength{\tabcolsep}{3pt}
	\begin{tabulary}{\textwidth}{L@{$\quad\quad$}L
			S[table-format=1.3]
			S[table-format=1.3]
			S[table-format=1.5]
			S[table-format=1.5]
			S[table-format=1.4]
			S[table-format=1.3]
		}
		\toprule
    \multirow{2}{*}{\bf\sffamily Model} & \multirow{2}{*}{\bf\sffamily Experiment} &
    \multicolumn{3}{c}{$\bm{\mathsf{Pr( t )}}$} &
    \multicolumn{3}{c}{$\bm{\mathsf{H(T)}}$} \\ \cmidrule(lr){3-5} \cmidrule(lr){6-8}
    & & {\bf\sffamily Unspecified} & {\bf\sffamily Specified} &
    {$\bm{\mathsf{ \left| \Delta \right|}}$} & {\bf\sffamily Unspecified} & {\bf\sffamily Specified} &
    {$\bm{\mathsf{\left| \Delta \right|}}$} \\ \midrule

DeepSeek & \multirow{2}{*}{\centering Bible} & 0.4 & 0.4 & 0 & 0.849 & 1.8 & 0.9 \\
GPT-4 &  & 0.4 & 0.2 & 0.2 & 2.19 & 3.22 & 1.03 \rule[-0.2cm]{0pt}{0pt} \\
DeepSeek & \multirow{2}{*}{\centering Bingo} & 0.872 & 0.7 & 0.2 & 0.715 & 0.8 & 0.1 \\
GPT-4 &  & 0.09 & 0.04 & 0.05 & 2.9 & 3.0 & 0.2 \rule[-0.2cm]{0pt}{0pt} \\
DeepSeek & \multirow{2}{*}{\centering Coin flip} & 1.0 & 0.998 & 0.002 & 0.00011 & 0.02 & 0.02 \\
GPT-4 &  & 1.0 & 0.999 & 0.00068 & 0.0002 & 0.0083 & 0.008 \rule[-0.2cm]{0pt}{0pt} \\
DeepSeek & \multirow{2}{*}{\centering Dart at map} & 0.783 & 0.6 & 0.2 & 0.987 & 1.0 & 0 \\
GPT-4 &  & 0.227 & 0.19 & 0.03 & 3.47 & 3.0 & 0.46 \rule[-0.2cm]{0pt}{0pt} \\
DeepSeek & \multirow{2}{*}{\centering Die roll} & 0.3 & 0.593 & 0.3 & 1.24 & 1.16 & 0.081 \\
GPT-4 &  & 0.96 & 0.924 & 0.0355 & 0.279 & 0.447 & 0.167 \rule[-0.2cm]{0pt}{0pt} \\
DeepSeek & \multirow{2}{*}{\centering Month and day} & 0.6 & 0.6 & 0.1 & 0.87 & 1.04 & 0.17 \\
GPT-4 &  & 0.36 & 0.028 & 0.34 & 2.85 & 3.8 & 0.9 \rule[-0.2cm]{0pt}{0pt} \\
DeepSeek & \multirow{2}{*}{\centering Playing cards} & 0.6 & 0.994 & 0.3 & 0.289 & 0.061 & 0.229 \\
GPT-4 &  & 0.4 & 0.13 & 0.26 & 2.44 & 3.49 & 1.05 \rule[-0.2cm]{0pt}{0pt} \\
DeepSeek & \multirow{2}{*}{\centering Rock-paper-scissors} & 0.7 & 0.978 & 0.3 & 0.5 & 0.15 & 0.4 \\
GPT-4 &  & 0.52 & 0.5 & 0 & 1.29 & 0.66 & 0.63 \rule[-0.2cm]{0pt}{0pt} \\
DeepSeek & \multirow{2}{*}{\centering Roulette} & 0.08 & 0.2 & 0.2 & 2.04 & 2.1 & 0 \\
GPT-4 &  & 0.13 & 0.16 & 0.02 & 3.39 & 3.34 & 0.06 \rule[-0.2cm]{0pt}{0pt} \\
DeepSeek & \multirow{2}{*}{\centering Shakespeare} & 0.979 & 0.4 & 0.6 & 0.144 & 1.1 & 1.0 \\
GPT-4 &  & 0.66 & 0.2 & 0.47 & 0.89 & 2.98 & 2.09 \rule[-0.2cm]{0pt}{0pt} \\

		\bottomrule
	\end{tabulary}
    \caption{Token probabilities and entropy over token probability distributions. The $\Delta$ columns show the difference between the unspecified and specified values. Values are reported up to 3 significant figures.}
    \label{table:results}
\end{table*}

GPT-4.1 generally generates responses with lower probability and higher entropy values when prompted with the probabilistic specification (fair or uniformly at random) compared to the unspecified prompt. The roulette scenario changes the probability from 0.13 to 0.16 and the entropy from 3.39 to 3.34 when the specification is included. Additionally, the specified results have lower entropy than the unspecified for the dart-on-map (3.47 to 3.00) and rock-paper-scissors (1.29 to 0.66) scenarios.

DeepSeek-Chat generated responses have increased probability values for 4 of the 10 scenarios when the prompt includes the probabilistic specification: die roll (0.3 to 0.593), playing cards(0.6 to 0.994), rock-paper-scissors (0.7 to 0.978), and roulette (0.08 to 0.2). Of these 4 scenarios, 3 produced lower entropy values when the prompt included the probabilistic specification: die roll (1.24 to 1.160), playing cards (0.289 to 0.061), and rock-paper-scissors (0.5 to 0.15). 

For both probability and entropy, DeepSeek-Chat and GPT-4.1 show the greatest response shift in the Shakespeare scenario. The explicit sampling strategy (uniformly at random) in the prompt decreases the response probability from 0.979 to 0.4 for DeepSeek-Chat and from 0.66 to 0.20 for GPT-4.1. Entropy increases from 0.144 to 1.1 for DeepSeek-Chat and from 0.89 to 2.98 for GPT-4.1.

\subsection{Generated to Theoretical Comparisons}

\begin{figure*}[t!]
\centering
\begin{subfigure}[b]{\textwidth}
    \centering
    \includegraphics[width=\columnwidth]{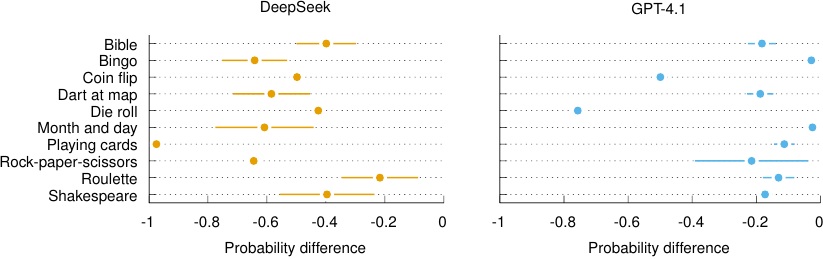}
    \caption{Difference between the selected token's probability and its theoretical value. The error bars show 1 SEM.}
    \label{fig:prob_diff}
\end{subfigure}
\begin{subfigure}[b]{\textwidth}
    \centering
    \includegraphics[width=\columnwidth]{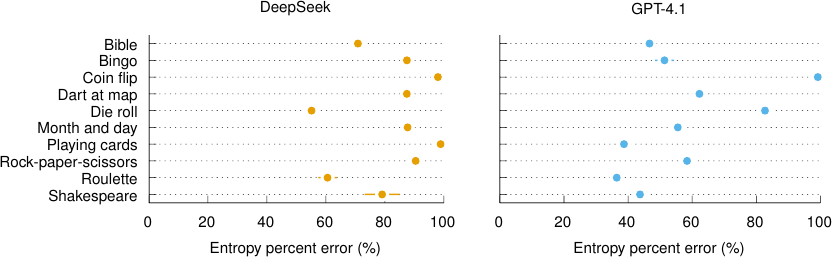}
    \caption{Percent error of each LLM's distribution over outputs with respect to the theoretical distribution. The error bars show 1 SEM.}
    \label{fig:ent_diff}
\end{subfigure}
    \caption{Comparisons of LLM token distributions to the theoretical distributions.}
    \label{bar}
\end{figure*}

Because the specified prompts produced results more closely aligned with the theoretical values, we use them as a basis for comparison. We compute the differences between the theoretical probability and entropy values and the corresponding mean values generated by the models.
%\autumn{this is bad obvs but should we write out explicity the difference or you think it's clear?}.
We display these comparisons in Figures \ref{fig:prob_diff} and \ref{fig:ent_diff}.

%\FloatBarrier

When explicitly prompted, GPT-4.1 produces probability and entropy values that are more closely aligned with the corresponding theoretical values than those generated by DeepSeek-Chat. For seven of the ten scenarios, GPT-4.1’s average probability differences are greater than -0.2, whereas all of DeepSeek-Chat’s probability differences fall below this threshold.

Both LLMs exhibit varying degrees of alignment with the theoretical values. GPT-4.1 achieves near-perfect probability alignment (for the selected token) in the Bingo (-0.03)  and Month-and-Day (-0.02) scenarios, whereas DeepSeek-Chat’s closest alignment occurs in the Roulette scenario (-0.22). GPT-4.1 shows its poorest probability alignment in the die roll (-0.75) and coin flip (-0.50) scenarios, which are two of the most elementary probability tasks in our scenario set. DeepSeek-Chat’s least aligned output occurs in the playing cards scenario (-0.97), always responding with 7 of hearts when given the specified prompt. 
Both models poorly capture entropy; for instance, the entropy associated with a coin flip deviates by nearly 100\%. GPT-4.1 exhibits lower percent error for most scenarios, but both models have higher than 30\% error modeling entropy in all scenarios.

%% file: latex/6_discussion.tex
\section{Discussion}
Because LLMs are considered effective tools for individual tasks or as components within larger processing pipelines, UQ is particularly important, not only to obtain confidence estimates, but also to understand precisely what those estimates represent. In some use cases, it may be necessary to expand UQ methods to capture task-specific requirements more comprehensively. In this work, we examine UQ in the context of prompts involving probabilistic scenarios, where an optimal model output would align with the corresponding theoretical probability distributions. Our experiments show that both DeepSeek-Chat and GPT-4.1 fail to achieve this alignment, specifically in straightforward cases such as a coin toss or die roll. 

Despite their failure to achieve probabilistic calibration, both DeepSeek-Chat and GPT-4.1 attained 100\% accuracy in response validity across all samples, demonstrating consistent in-domain knowledge of the probability-oriented tasks. However, their response certainty levels varied by scenario. For example, in the unspecified prompts, both DeepSeek-Chat and GPT-4.1 exhibited the highest certainty on the coin flip task (Pr($t$) $= 1.0$ and H($T$)~$\leq~0.0002$ for the token ``heads''). DeepSeek-Chat additionally produced a highly confident response in the Shakespeare scenario (Pr($t$) $= 0.979$ and H($T$) $= 0.144$ for the token ``Hamlet''). In contrast, the greatest response uncertainty was observed in the bingo (Pr($t$) $= 0.09$ and H($T$) $= 2.9$), dart on map (Pr($t$) $= 0.227$ and H($T$) $= 3.47$), and roulette (Pr($t$) $= 0.13$ and H($T$) $= 3.39$) scenarios for GPT-4.1, and in the die roll (Pr($t$) $= 0.3$; H($T$) $= 1.24$) and roulette (Pr($t$) $= 0.08$ and H($T$) $= 2.04$) scenarios for DeepSeek-Chat.

%Although GPT-4.1 and DeepSeek-Chat consistently produced valid in-domain responses, their outputs lacked probabilistic calibration. 
Our findings are consistent with \citet{lovering-etal-2025-language}, even with our difference in prompt style; we did not explicitly specify the constraints of the probabilistic scenario, but rather experimented with well-known scenarios. Additionally, our prompts elicited responses consisting solely of the final answer, ensuring that the observed biases are attributable exclusively to token-level probabilities. In typical UQ settings, certainty measurements must account for the auto-regressive generation process of LLMs; however, by removing this factor in our experiments, we demonstrate that systematic biases persist even in the absence of sequential generation effects. 

%Both models achieved 100\% accuracy in terms of response validity, indicating that they consistently demonstrated in-domain knowledge of the probability-oriented tasks. 

\begin{figure}[t]
\centering
    \centering
    \includegraphics[width=\columnwidth]{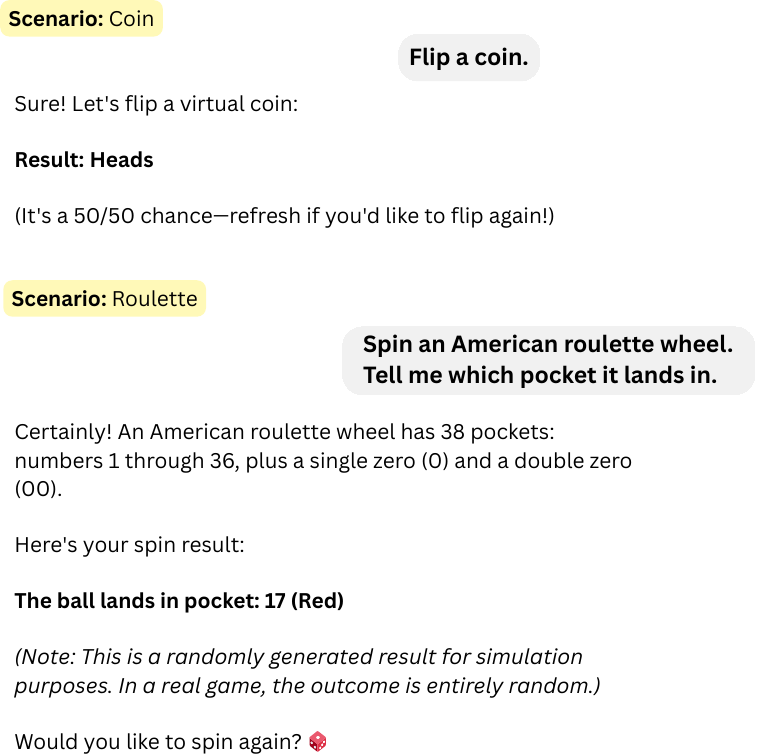}
    \caption{DeepSeek-Chat example dialogue for probabilistic reasoning about prompt scenarios.}
    \label{fig:deepseek_dialogue}
\end{figure}

\begin{figure}[t]
\centering
    \centering
    \includegraphics[width=\columnwidth]{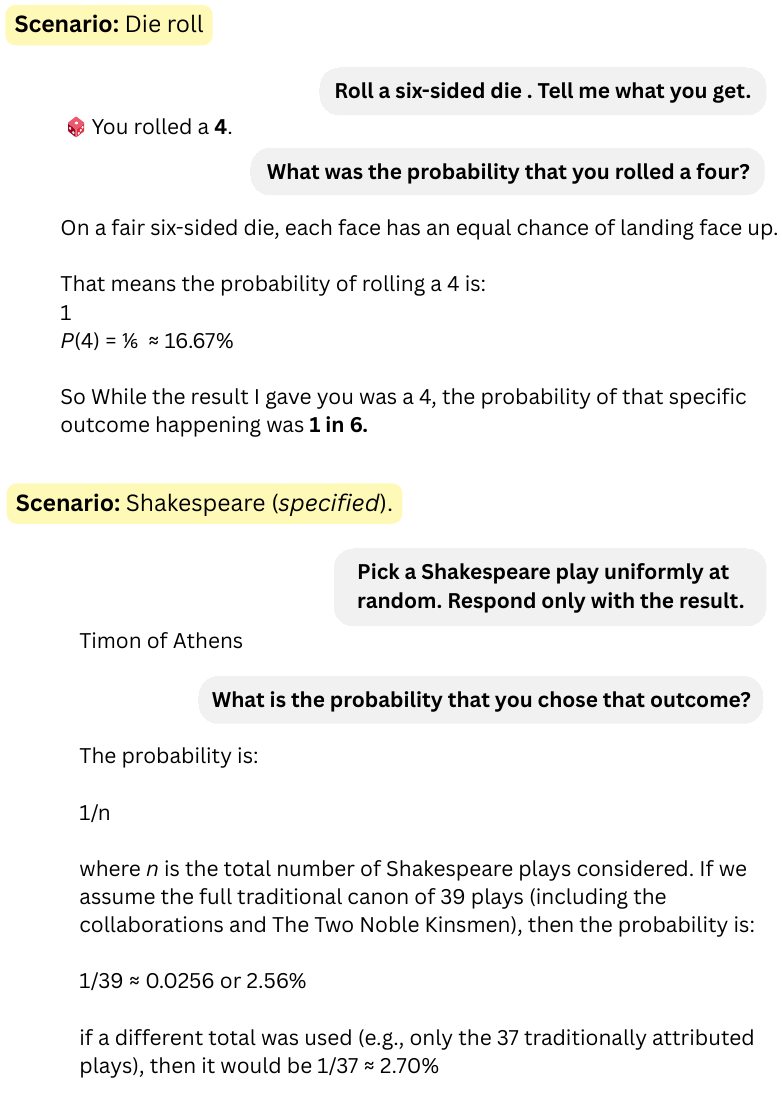}
    \caption{GPT-4.1 example dialogue for probabilistic reasoning about prompt scenarios.}
    \label{fig:gpt_dialogue}
\end{figure}

Motivated by our findings, we explore three additional questions for deeper examination and discussion:

\textbf{(1) Can LLMs accurately reason about the theoretical probabilities of our prompt scenarios?}
In our initial experimentation we found that DeepSeek-Chat provided its response along with its reasoning, as shown in the examples in Figure \ref{fig:deepseek_dialogue}. We conducted follow-up interactive chat sessions with GPT-4.1 to probe the model's reasoning processes and assess its understanding of the sampling techniques required by the statistical scenario prompts (examples shown in Figure \ref{fig:gpt_dialogue}). 

Both models demonstrate the ability to articulate the correct probability distributions for the prompted scenarios, reflecting accurate reasoning and in-domain knowledge. Despite their dialogue responses, DeepSeek-Chat and GPT-4.1 generated response tokens with probability and entropy values that did not align with the corresponding theoretical distributions. This discrepancy between \textit{verbalized reasoning} and \textit{token-level sampling} surfaces a critical gap between model reasoning explanations and probabilistic calibration.

\textbf{(2) Are LLMs appropriate and reliable tools for probability-oriented tasks where usability depends on alignment with theoretical distributions?} Our experimental results suggest that, while LLMs can produce valid outputs for probability-oriented tasks, they are not reliably aligned with the underlying theoretical distributions required for the desired performance. In applications where usability depends on accurate probabilistic calibration (e.g., simulations, randomized trial designs, decision-support systems involving chance) this misalignment could lead to systematic bias or misleading outcomes. The observed divergence between output certainty and theoretical probability indicates that LLMs, in their current form, may not be suitable as stand-alone tools for such tasks without additional calibration, fine-tuning, or post-processing to enforce distributional alignment. While our follow-up chatbot interactions suggest that LLMs possess the underlying knowledge to reason correctly about probabilities---giving the appearance that they are suitable for probability-oriented tasks---they are not inherently designed to generate outputs aligned with theoretical distributions. Instead, their outputs reflect the statistical patterns present in their training data. Thus, integrating these models into workflows that involves probabilistic behavior should require evaluation beyond traditional UQ prior to deployment.

\textbf{(3) How can uncertainty quantification methods be adapted to jointly evaluate response validity and distribution alignment for probability-driven tasks?} UQ is intended to provide a meaningful measure of a model’s response certainty, directly influencing user trust and perceived usability in a given task. While traditional UQ metrics can capture response validity (e.g., is the output valid for the prompted scenario constraints?), they do not account for distributional alignment (e.g., is the output sampled from the corresponding theoretical distribution required for an ``accurate'' response?). For probabilistic scenarios, we recommend that UQ be extended to either: (1) provide separate metrics: one for validity (compliance with task constraints) and one for probability alignment (closeness to the target distribution), or (2) define a composite metric: integrating both dimensions into a single score that reflects overall task suitability. In this way, UQ for probabilistic scenarios should provide insight into the ``certainty'' of distribution alignment.

%% file: latex/7_conclusion.tex
\section{Conclusion}

The divergence between certainty and probability has important implications for deploying LLMs in high-stakes decision-support contexts, where probabilistic calibration is critical for trustworthy system behavior. In this study, we examined the relationship between token-level certainty and theoretical probability alignment in LLMs, focusing on probabilistic scenarios with well-defined distributions. Across ten probability-oriented prompts, GPT-4.1 and DeepSeek-Chat consistently generated valid responses within scenario constraints; however, their token-level probability and entropy values differed from the corresponding theoretical distributions. These results highlight an important distinction in uncertainty quantification between response certainty and probabilistic calibration. Additional evaluation is required when alignment with a theoretical probability distribution is a critical aspect of the task. Without such assessment, a model’s apparent accuracy may mask deficiencies in its probabilistic calibration. 

%% file: 0_main.bbl
\begin{thebibliography}{14}
\providecommand{\natexlab}[1]{#1}

\bibitem[{Duan et~al.(2024)Duan, Cheng, Wang, Zavalny, Wang, Xu, Kailkhura, and
  Xu}]{duan-etal-2024-shifting}
Jinhao Duan, Hao Cheng, Shiqi Wang, Alex Zavalny, Chenan Wang, Renjing Xu,
  Bhavya Kailkhura, and Kaidi Xu. 2024.
\newblock \href {https://doi.org/10.18653/v1/2024.acl-long.276} {Shifting
  attention to relevance: Towards the predictive uncertainty quantification of
  free-form large language models}.
\newblock In \emph{Proceedings of the 62nd Annual Meeting of the Association
  for Computational Linguistics (Volume 1: Long Papers)}, pages 5050--5063,
  Bangkok, Thailand. Association for Computational Linguistics.

\bibitem[{Fadeeva et~al.(2024)Fadeeva, Rubashevskii, Shelmanov, Petrakov, Li,
  Mubarak, Tsymbalov, Kuzmin, Panchenko, Baldwin, Nakov, and
  Panov}]{fadeeva-etal-2024-fact}
Ekaterina Fadeeva, Aleksandr Rubashevskii, Artem Shelmanov, Sergey Petrakov,
  Haonan Li, Hamdy Mubarak, Evgenii Tsymbalov, Gleb Kuzmin, Alexander
  Panchenko, Timothy Baldwin, Preslav Nakov, and Maxim Panov. 2024.
\newblock \href {https://doi.org/10.18653/v1/2024.findings-acl.558}
  {Fact-checking the output of large language models via token-level
  uncertainty quantification}.
\newblock In \emph{Findings of the Association for Computational Linguistics:
  ACL 2024}, pages 9367--9385, Bangkok, Thailand. Association for Computational
  Linguistics.

\bibitem[{Gupta et~al.(2024)Gupta, Narasimhan, Jitkrittum, Rawat, Menon, and
  Kumar}]{gupta2024language}
Neha Gupta, Harikrishna Narasimhan, Wittawat Jitkrittum, Ankit~Singh Rawat,
  Aditya~Krishna Menon, and Sanjiv Kumar. 2024.
\newblock Language model cascades: Token-level uncertainty and beyond.
\newblock \emph{International Conference on Learning Representations}.

\bibitem[{Joshi et~al.(2017)Joshi, Choi, Weld, and
  Zettlemoyer}]{joshi-etal-2017-triviaqa}
Mandar Joshi, Eunsol Choi, Daniel Weld, and Luke Zettlemoyer. 2017.
\newblock \href {https://doi.org/10.18653/v1/P17-1147} {{T}rivia{QA}: A large
  scale distantly supervised challenge dataset for reading comprehension}.
\newblock In \emph{Proceedings of the 55th Annual Meeting of the Association
  for Computational Linguistics (Volume 1: Long Papers)}, pages 1601--1611,
  Vancouver, Canada. Association for Computational Linguistics.

\bibitem[{Kuhn et~al.(2023)Kuhn, Gal, and Farquhar}]{kuhn2023semantic}
Lorenz Kuhn, Yarin Gal, and Sebastian Farquhar. 2023.
\newblock Semantic uncertainty: Linguistic invariances for uncertainty
  estimation in natural language generation.
\newblock \emph{International Conference on Learning Representations}.

\bibitem[{Lin et~al.(2024)Lin, Trivedi, and Sun}]{lin-etal-2024-contextualized}
Zhen Lin, Shubhendu Trivedi, and Jimeng Sun. 2024.
\newblock \href {https://doi.org/10.18653/v1/2024.emnlp-main.578}
  {Contextualized sequence likelihood: Enhanced confidence scores for natural
  language generation}.
\newblock In \emph{Proceedings of the 2024 Conference on Empirical Methods in
  Natural Language Processing}, pages 10351--10368, Miami, Florida, USA.
  Association for Computational Linguistics.

\bibitem[{Liu et~al.(2025)Liu, Chen, Da, Chen, Lin, and
  Wei}]{liu2025uncertainty}
Xiaoou Liu, Tiejin Chen, Longchao Da, Chacha Chen, Zhen Lin, and Hua Wei. 2025.
\newblock Uncertainty quantification and confidence calibration in large
  language models: A survey.
\newblock In \emph{Proceedings of the 31st ACM SIGKDD Conference on Knowledge
  Discovery and Data Mining V. 2}, pages 6107--6117.

\bibitem[{Lovering et~al.(2025)Lovering, Krumdick, Lai, Reddy, Ebner, Kumar,
  Koncel-Kedziorski, and Tanner}]{lovering-etal-2025-language}
Charles Lovering, Michael Krumdick, Viet~Dac Lai, Varshini Reddy, Seth Ebner,
  Nilesh Kumar, Rik Koncel-Kedziorski, and Chris Tanner. 2025.
\newblock \href {https://doi.org/10.18653/v1/2025.acl-long.1417} {Language
  model probabilities are $not$ calibrated in numeric contexts}.
\newblock In \emph{Proceedings of the 63rd Annual Meeting of the Association
  for Computational Linguistics (Volume 1: Long Papers)}, pages 29218--29257,
  Vienna, Austria. Association for Computational Linguistics.

\bibitem[{Malinin and Gales(2021)}]{malinin2020uncertainty}
Andrey Malinin and Mark Gales. 2021.
\newblock Uncertainty estimation in autoregressive structured prediction.
\newblock \emph{International Conference on Learning Representations}.

\bibitem[{Reddy et~al.(2019)Reddy, Chen, and Manning}]{reddy2019coqa}
Siva Reddy, Danqi Chen, and Christopher~D Manning. 2019.
\newblock Coqa: A conversational question answering challenge.
\newblock \emph{Transactions of the Association for Computational Linguistics},
  7:249--266.

\bibitem[{Shorinwa et~al.(2025)Shorinwa, Mei, Lidard, Ren, and
  Majumdar}]{shorinwa2025survey}
Ola Shorinwa, Zhiting Mei, Justin Lidard, Allen~Z Ren, and Anirudha Majumdar.
  2025.
\newblock A survey on uncertainty quantification of large language models:
  Taxonomy, open research challenges, and future directions.
\newblock \emph{ACM Computing Surveys}.

\bibitem[{Vashurin et~al.(2025)Vashurin, Fadeeva, Vazhentsev, Rvanova, Vasilev,
  Tsvigun, Petrakov, Xing, Sadallah, Grishchenkov
  et~al.}]{vashurin2025benchmarking}
Roman Vashurin, Ekaterina Fadeeva, Artem Vazhentsev, Lyudmila Rvanova, Daniil
  Vasilev, Akim Tsvigun, Sergey Petrakov, Rui Xing, Abdelrahman Sadallah,
  Kirill Grishchenkov, and 1 others. 2025.
\newblock Benchmarking uncertainty quantification methods for large language
  models with lm-polygraph.
\newblock \emph{Transactions of the Association for Computational Linguistics},
  13:220--248.

\bibitem[{Xiong et~al.(2024)Xiong, Santilli, Kirchhof, Golinski, and
  Williamson}]{xiong2024efficient}
Miao Xiong, Andrea Santilli, Michael Kirchhof, Adam Golinski, and Sinead
  Williamson. 2024.
\newblock Efficient and effective uncertainty quantification for llms.
\newblock In \emph{Neurips Safe Generative AI Workshop 2024}.

\bibitem[{Zhang et~al.(2022)Zhang, Roller, Goyal, Artetxe, Chen, Chen, Dewan,
  Diab, Li, Lin et~al.}]{zhang2022opt}
Susan Zhang, Stephen Roller, Naman Goyal, Mikel Artetxe, Moya Chen, Shuohui
  Chen, Christopher Dewan, Mona Diab, Xian Li, Xi~Victoria Lin, and 1 others.
  2022.
\newblock Opt: Open pre-trained transformer language models.
\newblock \emph{arXiv preprint arXiv:2205.01068}.

\end{thebibliography}
